\documentclass[sigconf,nonacm]{acmart}
\setcopyright{none}

\usepackage[utf8]{inputenc}
\usepackage[T1]{fontenc}
\usepackage{url}
\usepackage{booktabs}
\usepackage{amsfonts}
\usepackage{nicefrac}
\usepackage{microtype}
\usepackage{xcolor}
\usepackage{graphicx}
\usepackage{subcaption}
\usepackage{float}
\usepackage{placeins}
\usepackage{dblfloatfix}
\usepackage{tabularx}
\usepackage{array}
\usepackage{adjustbox}

\hypersetup{
  hidelinks,
  pdfauthor={Aman Sunesh, Ali Alshehhi, Hivansh Dhakne},
  pdftitle={RequestRouter: Request-Boundary Routing for Efficient Single-GPU LLM Inference}
}

\newcommand{\system}{RequestRouter}

\title{\system{}: Request-Boundary Routing for Efficient Single-GPU LLM Inference}

\author{Aman Sunesh}
\affiliation{%
  \institution{New York University}
  \country{United States}}
\email{as18181@nyu.edu}

\author{Ali Alshehhi}
\affiliation{%
  \institution{New York University}
  \country{United States}}
\email{aa8148@nyu.edu}

\author{Hivansh Dhakne}
\affiliation{%
  \institution{New York University}
  \country{United States}}
\email{hd2296@nyu.edu}

\begin{document}

\begin{abstract}
\system{} is a lightweight request-boundary controller for reducing the latency and energy cost of single-GPU large language model inference. Rather than serving all requests with one static configuration, the system uses cheap request-level features to select one fixed inference mode per request, including FP16, quantized inference, speculative decoding, prefix caching, continuous batching, and hybrid modes such as GPTQ plus prefix caching and INT8 plus continuous batching. We evaluate \system{} using an 8B instruction-tuned language model served through vLLM on NVIDIA A100 GPUs. Across the full-scale A100 evaluation---26,500 fixed-mode evaluations followed by 3,500 online-controller evaluations, for 30,000 measured inference executions in total---the controller achieves a 2.10$\times$ mean latency speedup over FP16 and a 0.48$\times$ energy ratio on deployment-style workloads. A smaller matched evaluation with repeated measurements provides a controlled statistical check of this result: \system{} retains a 1.93$\times$ latency speedup (95\% CI: 1.88--1.98$\times$) and a 0.523 energy ratio (95\% CI: 0.506--0.540), showing that the gains persist under a more tightly controlled protocol. On a separate expanded automatic benchmark evaluation, the routed policy retains 99.6\% of FP16 macro accuracy. A 100,000-call CPU microbenchmark measures only 0.00475 ms mean routing overhead (0.00532 ms p99). Thus, simple request-aware routing can recover substantial serving efficiency without retraining or modifying the underlying LLM.
\end{abstract}


\maketitle

\section{Introduction}

Large language model inference is becoming a major source of computational cost in modern AI systems. Even when model weights are fixed, serving choices strongly affect latency, throughput, energy consumption, and memory pressure. This matters for low-carbon computing because inference workloads can run continuously at scale, and small per-request inefficiencies can accumulate into substantial operational energy cost.

Current serving deployments often use one static inference configuration across heterogeneous requests. This is convenient, but it ignores that different requests stress the GPU differently. Short interactive prompts are latency-sensitive, long-generation requests spend more time in decoding, shared-prefix chat workloads can benefit from cache reuse, and memory-pressure workloads may require more conservative modes. No single serving mode is best across all of these settings.

\system{} addresses this problem with a lightweight request-boundary controller. Before generation begins, the controller uses cheap request-level features such as prompt length, expected output length, shared-prefix status, memory pressure, batch pressure, and workload family to select one fixed inference mode. Candidate modes include FP16, GPTQ 4-bit quantization~\cite{gptq}, INT8 quantization, speculative decoding~\cite{speculative-decoding}, prefix caching, continuous batching, and hybrid configurations. The system does not retrain the model, modify the architecture, or switch modes inside a request.

The key observation is simple: energy-efficient serving does not require a single globally optimal inference configuration. Instead, many practical gains can be recovered by matching an existing inference optimization to the request structure that benefits from it. For example, prefix caching is useful only when requests share context, speculative decoding is most natural for decode-heavy generation, and quantized modes are useful when their efficiency gains do not introduce unacceptable quality changes. A request-boundary controller can exploit these differences without adding complex online scheduling logic.

This paper makes three contributions. First, we characterize how common LLM inference modes differ in latency and energy behavior across workload families. Second, we propose a low-overhead request-boundary controller for selecting among these modes. Third, we evaluate whether the resulting efficiency gains preserve benchmark quality and whether learned routers improve over a simple rule-based policy.

\section{Background and Motivation}

Modern LLM serving systems already include many optimizations for improving throughput and memory efficiency. Orca introduced iteration-level batching for transformer serving~\cite{yu2022orca}, while vLLM uses PagedAttention to manage KV-cache memory more efficiently~\cite{kwon2023efficient}. SGLang further improves structured LLM program execution and repeated-prefix reuse through mechanisms such as RadixAttention~\cite{zheng2024sglang}. These systems show that serving behavior is not determined only by the model architecture; the runtime and execution mode can substantially change the cost of inference.

Other work focuses on specific sources of efficiency. Quantization methods such as GPTQ~\cite{gptq} and AWQ~\cite{awq} reduce memory and compute cost by using lower-precision weights. Speculative decoding accelerates generation by using a draft model to propose tokens that are later verified by the target model~\cite{speculative-decoding}. Systems such as Sarathi-Serve and DistServe emphasize that prefill and decode phases create different bottlenecks and can benefit from phase-aware scheduling~\cite{agrawal2024sarathi,zhong2024distserve}. These techniques are complementary, but they are often evaluated as fixed modes or global serving configurations.

Request-aware routing has also been studied at the model-selection level. FrugalGPT uses learned cascades to choose among different LLM services, while RouteLLM learns to route queries between stronger and weaker models to trade serving cost against response quality~\cite{frugalgpt,routellm}. \system{} addresses a different control boundary: it keeps the target serving stack fixed and routes requests among \emph{execution modes} of that stack---for example quantized weights, speculative decoding, prefix reuse, and batching. Our focus is therefore not model selection, but low-overhead request-level selection of the inference mechanism used to execute a request.

From a low-carbon computing perspective, this creates an opportunity. If different inference modes are already available, then an energy-aware system does not always need to design a new model or deploy new hardware. It can instead choose the existing mode that best matches the request. This is especially attractive because it avoids the embodied and engineering costs of retraining, fine-tuning, or maintaining multiple task-specific model variants. The challenge is to make the routing decision cheaply enough that the controller does not erase the efficiency gains it is trying to recover.

\section{System and Evaluation}

\subsection{Inference Setting}

The full-scale evaluation uses a single NVIDIA A100 40 GB GPU, while the smaller matched repeated evaluation is conducted on an A100-SXM4-80GB. Both evaluations use Meta-Llama-3.1-8B-Instruct served through vLLM; results from the two configurations are reported separately rather than pooled. FP16 denotes IEEE 754 16-bit floating-point execution~\cite{ieee754} and is the unquantized reference configuration in our serving stack. Although FP16 itself uses less precision than FP32, here it serves as the baseline against which further compression and serving optimizations are measured. Optimized modes include INT8 and GPTQ 4-bit quantization, speculative decoding, prefix caching, continuous batching, and hybrid modes.

We focus on a single-GPU setting because it is common in small-scale research, prototyping, and constrained deployment environments. It is also a useful setting for isolating the effect of serving-mode selection without adding distributed scheduling, network communication, or multi-GPU placement as confounding factors. The goal is not to claim that one routing policy is universally optimal across all hardware, but to show that lightweight request-aware routing can recover meaningful efficiency even in a simple deployment setup.

\subsection{Workloads}

We evaluate two workload groups. Synthetic deployment-style workloads vary prompt and output length using short/short, short/long, long/short, and long/long request shapes. The short-prompt workloads approximate interactive or task-style requests, while the long-prompt workloads stress prefill and KV-cache behavior. Long-output workloads stress decode cost because the model must generate more tokens after the prompt has been processed.

We also include shared-prefix chat and memory-pressure long-context workloads. Shared-prefix chat represents settings where multiple requests reuse a common system prompt or conversation prefix. Memory-pressure workloads represent less ideal serving conditions where the GPU is not operating with abundant free memory. These workloads are important because they correspond to deployment behaviors that are not captured by a simple short-prompt benchmark.

Automatic benchmark workloads include MMLU-Pro~\cite{mmlu-pro}, GSM8K~\cite{gsm8k}, TruthfulQA~\cite{truthfulqa}, GPQA~\cite{gpqa}, and MLU~\cite{mlu}. Synthetic workloads measure deployment efficiency, while automatic benchmarks act as a quality gate. This separation is important: a mode can look efficient on latency and energy while still being undesirable if it meaningfully changes output behavior.

\subsection{Metrics}

For each mode and workload, we measure total latency, throughput, energy per generated token, peak GPU memory, and output quality. Energy is estimated by polling GPU power with NVML every 50 ms, integrating power over time, and dividing by generated tokens. Latency speedup is FP16 latency divided by mode latency, so higher is better. Energy ratio is mode energy per token divided by FP16 energy per token, so lower is better. Accuracy deltas are measured in percentage points relative to FP16.

For the smaller matched evaluation, each synthetic workload--mode pair is measured five times using identical requests across modes. Confidence intervals are computed by bootstrapping workload variants as clusters, keeping repeated measurements of the same workload together. We report 95\% confidence intervals where space permits.

Separately, for quality, we use 300 paired examples per benchmark with deterministic decoding. MMLU-Pro, TruthfulQA, GPQA, and MLU use multiple-choice accuracy, comparing the extracted answer label with the ground-truth label. GSM8K uses normalized final-answer exact match with numeric normalization. The same examples are evaluated across all serving modes.

Energy per token is the main low-footprint metric in this study. It does not directly measure carbon emissions because carbon intensity depends on grid mix, datacenter location, and time of execution. However, reducing energy per generated token is still a direct systems-level objective and a necessary component of lower-carbon inference. We therefore report energy ratios relative to FP16 rather than making absolute carbon claims.

\subsection{Request-Boundary Controller}

The online controller selects one fixed inference mode per request before generation. It classifies requests as batched, shared-prefix, memory-pressure, prefill-heavy, or decode-heavy. Table~\ref{tab:routing-policy} summarizes the rule-based routing policy. The policy sends batched requests to INT8 plus continuous batching, shared-prefix requests to GPTQ plus prefix caching, memory-pressure and several synthetic workloads to GPTQ 4-bit, long-output decode-heavy requests to speculative decoding, and multiple-choice benchmark-style requests to INT8 quantization. FP16 remains an emergency fallback.

\begin{table}[t]
\centering
\caption{Rule-based routing policy used by \system{}.}
\label{tab:routing-policy}
\footnotesize
\setlength{\tabcolsep}{4pt}
\renewcommand{\arraystretch}{1.08}
\begin{tabularx}{\columnwidth}{@{}>{\raggedright\arraybackslash}X>{\raggedright\arraybackslash}X@{}}
\toprule
Request condition & Selected mode \\
\midrule
Batched / multi-request serving & INT8 + continuous batching \\
Shared-prefix chat & GPTQ + prefix caching \\
Memory-pressure long context & GPTQ 4-bit \\
Short/short, long/short, long/long synthetic & GPTQ 4-bit \\
Decode-heavy long-output generation & Speculative decoding \\
Benchmark-style multiple-choice request & INT8 quantization \\
Fallback / unsupported request type & FP16 or INT8 \\
\bottomrule
\end{tabularx}
\end{table}

The rule-based policy is intentionally simple. Its goal is not to learn a complex control strategy, but to test whether request structure alone is enough to recover energy and latency gains from existing inference modes. A same-environment CPU microbenchmark over 100,000 routing decisions measures 0.00475 ms mean controller latency (SD: 0.00092 ms; p95: 0.00500 ms; p99: 0.00532 ms), negligible relative to end-to-end generation latency.

\section{Results}

\subsection{Fixed-Mode Characterization}

Different inference modes dominate different workload families. GPTQ 4-bit provides strong latency and energy gains on many synthetic workloads, while INT8 is more competitive on several benchmark-style workloads. Speculative decoding is particularly effective for decode-heavy, long-output requests, whereas continuous batching is better suited to batched or multi-request serving with higher request pressure. Prefix caching is most useful when requests share substantial repeated context. This variation supports the central design choice of \system{}: routing should depend on workload structure rather than relying on one static serving configuration.

Hybrid modes further show that optimizations can be complementary. GPTQ plus prefix caching reduces latency from 1903 ms to 942 ms on shared-prefix workloads and lowers energy from 3.26 to 1.36 J/token. INT8 plus continuous batching reduces multi-request latency from 1840 ms to 1361 ms and lowers energy from 1.32 to 0.83 J/token.

\subsection{Online Controller Efficiency}

The full-scale A100 evaluation comprises 26,500 fixed-mode inference evaluations and 3,500 additional online-controller evaluations, for 30,000 measured inference executions in total. The fixed-mode sweep consists of 24,000 synthetic evaluations (six workload families, 100 variants, eight modes, and five trials) and 2,500 benchmark evaluations (five benchmark families, 100 prompts, five modes, and one trial). The online-controller evaluation adds 3,000 synthetic and 500 benchmark executions. On the deployment-style workloads, \system{} achieves a 2.10$\times$ mean latency speedup over FP16 and a 0.48$\times$ energy/token ratio.

To assess whether these gains persist under a more tightly controlled statistical protocol, we additionally evaluate 144 synthetic workload variants across six serving modes with five matched repetitions per workload--mode pair (4,320 successful generations). Despite using only roughly one-seventh as many generations as the full-scale evaluation, \system{} retains a 1.926$\times$ latency speedup over FP16 (95\% workload-cluster bootstrap CI: 1.875--1.978$\times$) and a 0.523 energy/token ratio (95\% CI: 0.506--0.540). Latency and energy both improve relative to FP16 in all six synthetic workload families.

The largest gains occur on deployment-style workloads where request structure provides useful routing signals. Shared-prefix workloads benefit from prefix reuse, while decode-heavy requests can benefit from speculative decoding. This variation across workload families motivates request-aware selection rather than deploying one inference mode globally.

\begin{figure}[t]
    \centering
    \includegraphics[width=\columnwidth]{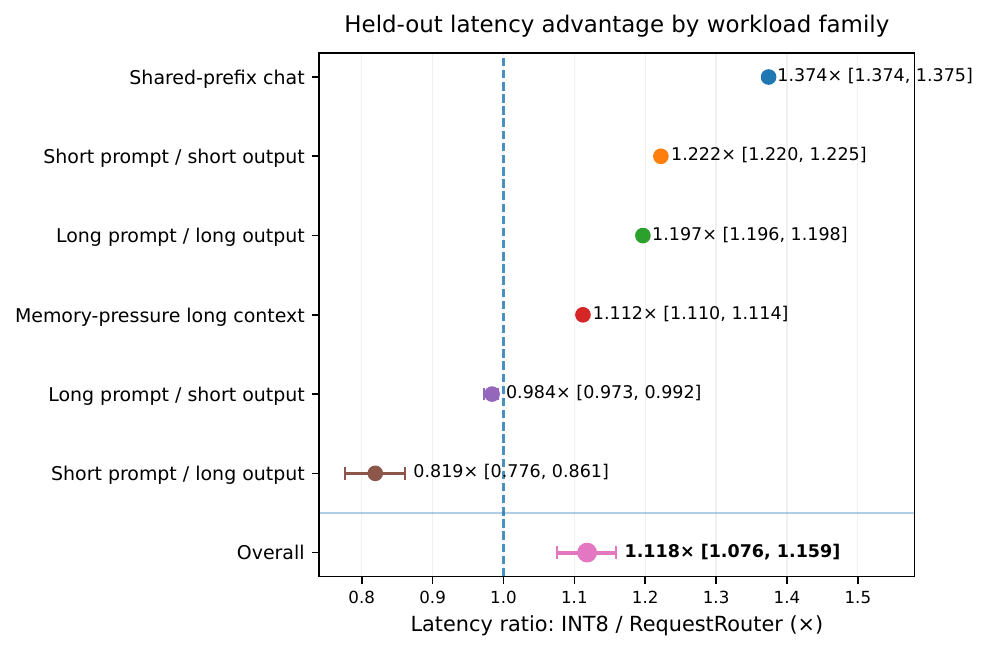}
    \caption{Held-out latency comparison between \system{} and the frozen
    quality-eligible static policy selected on calibration data (INT8).
    Values above 1 indicate lower latency for \system{}. Whiskers show
    95\% workload-cluster bootstrap confidence intervals. \system{} is
    faster in four of six workload families and achieves an overall
    1.118$\times$ latency advantage (95\% CI:
    1.076--1.159$\times$).}
    \label{fig:heldout-family}
\end{figure}

To test whether routing is preferable to a fixed optimized configuration, we split the 144 workload variants into 72 calibration and 72 held-out variants. Using the 1.5-percentage-point quality criterion described below, INT8 is the best quality-eligible static mode selected on the calibration set. On the held-out set, \system{} reduces mean latency from 790.8 to 726.8 ms. The paired workload-level latency advantage is 1.118$\times$ (95\% CI: 1.076--1.159$\times$), with \system{} faster on 66.7\% of held-out workload variants. Figure~\ref{fig:heldout-family} shows the family-level latency advantage of \system{} over the best quality-eligible static policy selected on the calibration set (INT8), together with 95\% confidence intervals.

\subsection{Quality Preservation}

Energy savings are only useful if routing does not meaningfully degrade model behavior. Table~\ref{tab:auto-accuracy} reports automatic benchmark accuracy by mode. The results show why quality-aware routing is necessary: GPTQ 4-bit is fast and energy-efficient but can reduce accuracy on some benchmarks, while INT8, speculative decoding, and prefix caching stay closer to FP16 on several tasks.

\begin{table}[t]
\centering
\caption{Automatic benchmark accuracy on 300 paired examples per benchmark. Multiple-choice benchmarks use answer-label accuracy; GSM8K uses normalized final-answer exact match. Values are percentages.}
\label{tab:auto-accuracy}
\footnotesize
\setlength{\tabcolsep}{3pt}
\renewcommand{\arraystretch}{1.08}
\begin{tabular*}{\columnwidth}{@{\extracolsep{\fill}}lccccc@{}}
\toprule
Mode & MMLU-Pro & GSM8K & TQA & GPQA & MLU \\
\midrule
FP16 & 36.3 & 83.3 & 54.0 & 29.0 & 50.3 \\
GPTQ 4-bit & 35.0 & 80.3 & 53.7 & 30.0 & 46.7 \\
INT8 & 35.7 & 84.0 & 55.7 & 28.3 & 49.0 \\
Spec. dec. & 36.3 & 83.3 & 53.7 & 29.7 & 50.3 \\
Prefix cache & 36.3 & 83.3 & 53.7 & 29.7 & 50.3 \\
\system{} & 35.7 & 83.3 & 55.7 & 28.3 & 49.0 \\
\bottomrule
\end{tabular*}
\end{table}

We define a serving mode as quality-eligible when no benchmark accuracy falls by more than 1.5 percentage points relative to FP16. Under the actual routed policy, macro accuracy is 50.4\%, compared with 50.6\% for FP16, corresponding to 99.6\% quality retention. The worst per-benchmark routed delta is $-1.33$ percentage points and remains inside the 1.5-point quality gate. By contrast, always-GPTQ reaches a worst-case delta of $-3.67$ points. This illustrates why the fastest unconstrained mode is not necessarily an admissible static deployment policy.

\subsection{Rule-Based and Learned Routing}

\begin{figure}[t]
    \centering
    \includegraphics[width=\columnwidth]{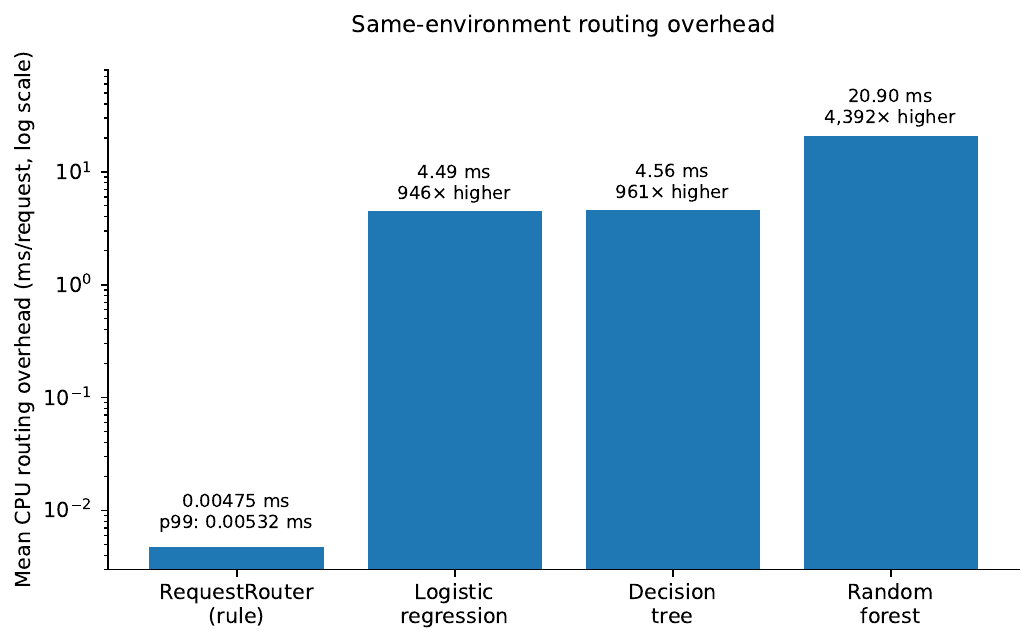}
    \caption{Same-environment CPU routing overhead. Across 100,000 routing  decisions, the rule controller requires 0.00475 ms/request on average  (SD: 0.00092 ms; p95: 0.00500 ms; p99: 0.00532 ms), compared with 4.49 ms for logistic regression, 4.56 ms for a decision tree, and 20.90 ms for a random forest. The logarithmic scale highlights the orders-of-magnitude difference in routing cost.}
    \label{fig:routing-overhead}
\end{figure}

We also trained decision-tree, random-forest, and logistic-regression routers to imitate a constraint-aware oracle. The oracle selects the fastest measured mode satisfying quality, energy, and memory constraints. Table~\ref{tab:learned-summary} summarizes the policy comparison on the learned-routing dataset. Although random forest reaches the highest oracle-label match rate, learned routing remains substantially more expensive. Figure~\ref{fig:routing-overhead} shows that the rule controller is orders of magnitude cheaper than the learned routing pipelines in the same execution environment. The simple rule controller is therefore the strongest practical policy in this setting: it reaches 2.38$\times$ speedup and a 0.47$\times$ energy ratio, close to the oracle's 2.41$\times$ speedup and 0.45$\times$ energy ratio, while same-environment timing measures only 0.00475 ms mean routing overhead. The rule-controller p99 latency is only 0.00532 ms, showing that its low mean overhead is not masking a large routing-latency tail.

\begin{table}[t]
\centering
\caption{Policy outcomes on the learned-routing dataset. CPU overhead is remeasured in the same execution environment and reported in ms/request.}
\label{tab:learned-summary}
\footnotesize
\setlength{\tabcolsep}{3pt}
\renewcommand{\arraystretch}{1.08}
\begin{tabular*}{\columnwidth}{@{\extracolsep{\fill}}lcccc@{}}
\toprule
Policy & Speedup & Energy & Match & Overhead \\
\midrule
FP16 & 1.00$\times$ & 1.00$\times$ & 0.2\% & 0.0000 \\
Oracle & 2.41$\times$ & 0.45$\times$ & 100.0\% & 0.0000 \\
Rule & 2.38$\times$ & 0.47$\times$ & 54.4\% & 0.0048 \\
LogReg & 2.26$\times$ & 0.46$\times$ & 51.1\% & 4.49 \\
Tree & 2.26$\times$ & 0.46$\times$ & 51.1\% & 4.56 \\
Forest & 2.22--2.23$\times$ & 0.44$\times$ & 60.0\% & 20.9 \\
\bottomrule
\end{tabular*}
\end{table}

This result is important for low-carbon computing. A controller that saves GPU energy but adds substantial CPU overhead or complexity may be unattractive in practice. \system{} instead shows that lightweight workload rules can capture much of the available efficiency without turning the router itself into a costly model.

\paragraph{Hardware transfer.}
The A100 remains the primary controlled platform, but we also performed smaller transfer checks on newer accelerators. On an NVIDIA L40S, \system{} achieves a $2.05\pm0.48\times$ mean latency speedup and a $0.53\pm0.22$ energy/token ratio across 305 matched measurements. On an RTX PRO 6000 Blackwell, the supported-mode evaluation achieves $2.06\pm0.27\times$ speedup and a $0.40\pm0.08$ energy/token ratio across 205 measurements. These smaller checks are not pooled with the main A100 experiment but indicate that the effect is not specific to the older accelerator.

\subsection{Low-Carbon Interpretation}

\system{} directly reduces measured GPU energy per generated token; it does not by itself establish a carbon-emissions reduction. If $E_{\mathrm{tok}}$ is GPU energy in J/token, $c$ is grid carbon intensity in gCO$_2$e/kWh, and $p$ is datacenter PUE, an operational estimate is
\[
C_{\mathrm{op}} =
\frac{E_{\mathrm{tok}}}{3.6\times10^6}\,c\,p .
\]

Thus, when two modes run at the same location, time, and PUE, their operational-emissions ratio equals their energy/token ratio. The measured 0.48 energy ratio therefore corresponds to the same relative operational-emissions reduction under low-, medium-, or high-carbon electricity (e.g., 50, 200, or 500 gCO$_2$e/kWh), while the absolute CO$_2$e savings scale linearly with carbon intensity.

We do not claim a measured embodied-carbon reduction. Within each evaluation, all compared modes execute on the same physical GPU, and allocating its manufacturing footprint per generated token would require assumptions about hardware lifetime, utilization, replacement, and total demand. Accordingly, our result should be interpreted as an operational energy-efficiency result rather than a complete life-cycle assessment.

\section{Discussion}

\system{} suggests that low-footprint LLM inference does not require a single universally efficient serving mode. Instead, practical gains can come from matching existing serving optimizations to request structure. This framing is useful because quantization, speculative decoding, prefix caching, and batching are often treated as separate techniques, while \system{} treats them as selectable operating modes within one serving stack.

The results also highlight the importance of controller overhead. Learned routers may be appealing because they can imitate an oracle, but their deployment value depends on the full cost of making the routing decision. In our experiments, the rule-based controller is competitive because it captures the main structural cases while adding almost no CPU overhead. This suggests that future learned controllers should optimize not only oracle-label accuracy, but also end-to-end latency, energy, and routing cost.

\section{Limitations and Future Work}

A limitation is that the controller operates only at request boundaries and assumes that the selected serving mode is warm or otherwise available. Cold reconstruction of distinct vLLM engines is not viable per request (measured loads are approximately 89--103 s); pairwise tests show
that some modes can co-reside, while tested speculative-decoding combinations cannot. Practical deployment therefore requires resident compatible engines or coarser-grained transitions. Finer-grained prefill/decode adaptation remains future work.

The synthetic workloads are structured rather than collected from a production trace. They are useful because they isolate prompt length, output length, shared-prefix behavior, and memory pressure, but real deployments may contain messier mixtures of these factors. Future work should evaluate request-boundary routing on traces from interactive chat, coding assistants, retrieval-augmented generation, and batch inference.

Efficiency is also not equivalent to sustainability. Lower latency and energy per request can reduce marginal serving cost and thereby induce rebound effects: if efficiency leads to more AI requests or broader deployment, total energy use and emissions can still increase~\cite{luccioni2024powerhungry}. We therefore make no claim that the per-token savings measured here necessarily reduce system-wide or economy-wide emissions.

A final direction is to connect request-aware inference modes with grid-aware or carbon-aware placement. \system{} reduces energy per generated token on a single GPU. A larger system could combine this with decisions about when and where to run inference based on electricity carbon intensity, latency requirements, and queueing constraints.

\paragraph{Artifacts.}
The controller implementation, workload definitions, benchmarking scripts, and analysis code are publicly available at \url{https://github.com/ModeSwitch-LLM/ModeSwitch-LLM}.

\section{Conclusion}

\system{} is a lightweight request-boundary controller for efficient single-GPU LLM inference. Across a full-scale evaluation campaign containing 26,500 fixed-mode and 3,500 online-controller evaluations (30,000 measured inference executions total), the system achieves a 2.10$\times$ mean latency speedup over FP16 on deployment-style workloads while reducing energy per generated token to 0.48$\times$ the FP16 baseline. A separate matched repeated evaluation of 4,320 generations yields a 1.93$\times$ latency speedup (95\% CI: 1.88--1.98$\times$) and a 0.523 energy ratio (95\% CI: 0.506--0.540).

On 72 held-out workload variants, \system{} also outperforms the frozen best quality-eligible static policy with a paired 1.118$\times$ latency advantage (95\% CI: 1.076--1.159$\times$), while retaining 99.6\% of FP16 macro benchmark accuracy. Mean routing overhead is only 0.00475 ms/request.

The broader result is that existing inference optimizations need not be treated as one global configuration: inexpensive request-aware selection can recover substantial latency and energy efficiency while respecting quality constraints. Such routing complements, rather than replaces, larger-scale carbon-aware scheduling and placement.

\bibliographystyle{ACM-Reference-Format}
\bibliography{references}

@article{mmlu-pro,
  title   = {{MMLU-Pro}: A More Robust and Challenging Multi-Task
             Language Understanding Benchmark},
  author  = {Wang, Yubo and Ma, Xueguang and Zhang, Ge and Ni, Yuansheng
             and Chandra, Abhranil and Guo, Shiguang and Ren, Weiming
             and Pan, Aaran and Zhang, Yuntian and Xu, Pengfei and others},
  journal = {arXiv preprint arXiv:2406.01574},
  year    = {2024}
}

@article{gsm8k,
  title   = {Training Verifiers to Solve Math Word Problems},
  author  = {Cobbe, Karl and Kosaraju, Vineet and Bavarian, Mohammad
             and Chen, Mark and Jun, Heewoo and Kaiser, Lukasz
             and Plappert, Matthias and Tworek, Jerry and Hilton, Jacob
             and Nakano, Reiichiro and Hesse, Christopher and Schulman, John},
  journal = {arXiv preprint arXiv:2110.14168},
  year    = {2021}
}

@article{truthfulqa,
  title   = {{TruthfulQA}: Measuring How Models Mimic Human Falsehoods},
  author  = {Lin, Stephanie and Hilton, Jacob and Evans, Owain},
  journal = {arXiv preprint arXiv:2109.07958},
  year    = {2021}
}

@article{gpqa,
  title   = {{GPQA}: A Graduate-Level {Google}-Proof {Q\&A} Benchmark},
  author  = {Rein, David and Hou, Betty Li and Stickland, Asa Cooper
             and Petty, Jackson and Pang, Richard Yuanzhe
             and Dirani, Julien and Michael, Julian and Bowman, Samuel R.},
  journal = {arXiv preprint arXiv:2311.12022},
  year    = {2023}
}

@article{mlu,
  title   = {Measuring Massive Multitask Language Understanding},
  author  = {Hendrycks, Dan and Burns, Collin and Basart, Steven
             and Zou, Andy and Mazeika, Mantas and Song, Dawn
             and Steinhardt, Jacob},
  journal = {arXiv preprint arXiv:2009.03300},
  year    = {2020}
}

@article{gptq,
  title   = {{GPTQ}: Accurate Post-Training Quantization for Generative
             Pre-Trained Transformers},
  author  = {Frantar, Elias and Ashkboos, Saleh and Hoefler, Torsten
             and Alistarh, Dan},
  journal = {arXiv preprint arXiv:2210.17323},
  year    = {2022}
}

@article{awq,
  title   = {{AWQ}: Activation-Aware Weight Quantization for {LLM}
             Compression and Acceleration},
  author  = {Lin, Ji and Tang, Jiaming and Tang, Haotian and Yang, Shang
             and Dang, Xingyu and Han, Song},
  journal = {arXiv preprint arXiv:2306.00978},
  year    = {2023}
}

@article{routellm,
  title={RouteLLM: Learning to Route LLMs with Preference Data},
  author={Ong, Isaac and Almahairi, Amjad and Wu, Vincent and
          Chiang, Wei-Lin and Wu, Tianhao and Gonzalez, Joseph E.
          and Kadous, M. Waleed and Stoica, Ion},
  journal={arXiv preprint arXiv:2406.18665},
  year={2024}
}

@article{frugalgpt,
  title={FrugalGPT: How to Use Large Language Models While Reducing Cost and Improving Performance},
  author={Chen, Lingjiao and Zaharia, Matei and Zou, James},
  journal={arXiv preprint arXiv:2305.05176},
  year={2023}
}

@inproceedings{luccioni2024powerhungry,
  title={Power Hungry Processing: Watts Driving the Cost of AI Deployment?},
  author={Luccioni, Alexandra Sasha and Jernite, Yacine and Strubell, Emma},
  booktitle={Proceedings of the 2024 ACM Conference on Fairness, Accountability, and Transparency},
  pages={85--99},
  year={2024},
  doi={10.1145/3630106.3658542}
}

@misc{ieee754,
  author={{IEEE}},
  title={{IEEE Standard for Floating-Point Arithmetic, IEEE Std 754-2019}},
  year={2019}
}

@inproceedings{speculative-decoding,
  title     = {Fast Inference from Transformers via Speculative Decoding},
  author    = {Leviathan, Yaniv and Kalman, Matan and Matias, Yossi},
  booktitle = {International Conference on Machine Learning},
  pages     = {19274--19286},
  year      = {2023},
  organization = {PMLR}
}

@inproceedings{agrawal2024sarathi,
  author    = {Amey Agrawal and Nitin Kedia and Ashish Panwar and
               Jayashree Mohan and Nipun Kwatra and Bhargav S. Gulavani
               and Alexey Tumanov and Ramachandran Ramjee},
  title     = {Taming Throughput-Latency Tradeoff in {LLM} Inference
               with {Sarathi-Serve}},
  booktitle = {18th USENIX Symposium on Operating Systems Design and
               Implementation (OSDI 24)},
  pages     = {117--134},
  publisher = {USENIX Association},
  year      = {2024}
}

@inproceedings{kwon2023efficient,
  author    = {Woosuk Kwon and Zhuohan Li and Siyuan Zhuang and
               Ying Sheng and Lianmin Zheng and Cody Hao Yu and
               Joseph E. Gonzalez and Hao Zhang and Ion Stoica},
  title     = {Efficient Memory Management for Large Language Model
               Serving with {PagedAttention}},
  booktitle = {Proceedings of the 29th ACM Symposium on Operating
               Systems Principles (SOSP '23)},
  publisher = {ACM},
  year      = {2023}
}

@inproceedings{yu2022orca,
  author    = {Gyeong-In Yu and Joo Seong Jeong and Geon-Woo Kim and
               Soojeong Kim and Byung-Gon Chun},
  title     = {Orca: A Distributed Serving System for
               Transformer-Based Generative Models},
  booktitle = {16th USENIX Symposium on Operating Systems Design and
               Implementation (OSDI 22)},
  pages     = {521--538},
  publisher = {USENIX Association},
  year      = {2022}
}

@inproceedings{zheng2024sglang,
  author    = {Lianmin Zheng and Liangsheng Yin and Zhiqiang Xie and
               Chuyue Sun and Jeff Huang and Cody Hao Yu and Shiyi Cao
               and Christos Kozyrakis and Ion Stoica and
               Joseph E. Gonzalez and Clark Barrett and Ying Sheng},
  title     = {{SGLang}: Efficient Execution of Structured Language
               Model Programs},
  booktitle = {Advances in Neural Information Processing Systems 37
               (NeurIPS 2024)},
  pages     = {62557--62583},
  year      = {2024}
}

@inproceedings{zhong2024distserve,
  author    = {Yinmin Zhong and Shengyu Liu and Junda Chen and
               Jianbo Hu and Yibo Zhu and Xuanzhe Liu and Xin Jin and
               Hao Zhang},
  title     = {{DistServe}: Disaggregating Prefill and Decoding for
               Goodput-Optimized Large Language Model Serving},
  booktitle = {18th USENIX Symposium on Operating Systems Design and
               Implementation (OSDI 24)},
  pages     = {193--210},
  publisher = {USENIX Association},
  year      = {2024}
}

\end{document}